\pdfoutput=1

\documentclass[11pt]{article}

\usepackage[]{acl}

\usepackage{times}
\usepackage{latexsym}

\usepackage[T1]{fontenc}

\usepackage[utf8]{inputenc}

\usepackage{microtype}

\usepackage{inconsolata}

\usepackage{graphicx}

\usepackage{times}
\usepackage{latexsym}

\usepackage[normalem]{ulem}
\usepackage{mathrsfs}
\usepackage{amsthm}
\usepackage{amsmath}
\usepackage{amssymb, bm}
\usepackage{multirow}
\usepackage{booktabs}
\usepackage{enumitem}
\usepackage{times}
\usepackage{latexsym}
\usepackage{array}
\usepackage{multirow}
\usepackage{graphicx}
\usepackage{amsfonts}
\usepackage{subcaption}
\usepackage{xspace}
\usepackage{tabularx}
\usepackage{algorithm}
\usepackage{arydshln}
\usepackage{algpseudocode}
\usepackage{colortbl}
\usepackage{tcolorbox}
\usepackage{makecell}

\definecolor{mypurple}{RGB}{111,61,121}
\definecolor{myblue}{RGB}{46,88,180}
\definecolor{myred}{RGB}{181,68,106}
\definecolor{textorange}{RGB}{237,125,49}
\definecolor{textblue}{RGB}{46,117,181}
\definecolor{textgreen}{RGB}{112,173,71}

\usepackage{xspace}
 \usepackage{amsfonts,amssymb}
\newcommand{\ours}{LENS\xspace}

\title{Enhancing Lexicon-Based Text Embeddings with Large Language Models}

\author{Yibin Lei\textsuperscript{1}, Tao Shen\textsuperscript{2}, Yu Cao\textsuperscript{3}, Andrew Yates\textsuperscript{1,4} \\
\textsuperscript{1}{University of Amsterdam} \quad \textsuperscript{2}{University of Technology Sydney} \\
\textsuperscript{3}{Tencent IEG}
\quad \textsuperscript{4}{Johns Hopkins University, HLTCOE} \\
\texttt{y.lei@uva.nl}, \texttt{tao.shen@uts.edu.au}\\
\texttt{rainyucao@tencent.com},
\texttt{andrew.yates@jhu.edu}
}

\begin{document}
\maketitle
\begin{abstract}
Recent large language models (LLMs) have demonstrated exceptional performance on general-purpose text embedding tasks. While dense embeddings have dominated related research, we introduce the first lexicon-based embeddings (\ours) leveraging LLMs that achieve competitive performance on these tasks.
\ours consolidates the vocabulary space through token embedding clustering to handle the issue of token redundancy in LLM vocabularies. To further improve performance, we investigate bidirectional attention and various pooling strategies.
Specifically, \ours simplifies lexical matching with redundant vocabularies by assigning each dimension to a specific token cluster, where semantically similar tokens are grouped together.
Extensive experiments demonstrate that \ours outperforms dense embeddings on the Massive Text Embedding Benchmark (MTEB), delivering compact representations with dimensionality comparable to dense counterparts.
Furthermore, \ours inherently supports efficient embedding dimension pruning without any specialized objectives like Matryoshka Representation Learning.
Notably, combining \ours with dense embeddings achieves state-of-the-art performance on the retrieval subset of MTEB (i.e., BEIR).\footnote{Our code and models are available at \url{https://github.com/Yibin-Lei/LENS}.}

\end{abstract}

\section{Introduction}
Text embeddings are vector representations of text that power a wide range of applications, including retrieval, question answering, semantic textual similarity, and clustering. Recent advances in LLMs have shown that a single model can generate embeddings excelling across diverse tasks, highlighting their versatility~\citep{e5_mistral, bge_en_icl, llm2vec, lei-etal-2024-meta, grit_lm, SFR-embedding-2, nv_embed}.

\begin{figure}[t]
    \centering
    \setlength{\abovecaptionskip}{0.3cm}
    \setlength{\belowcaptionskip}{-0.5cm}
    \includegraphics[width=0.95\linewidth]{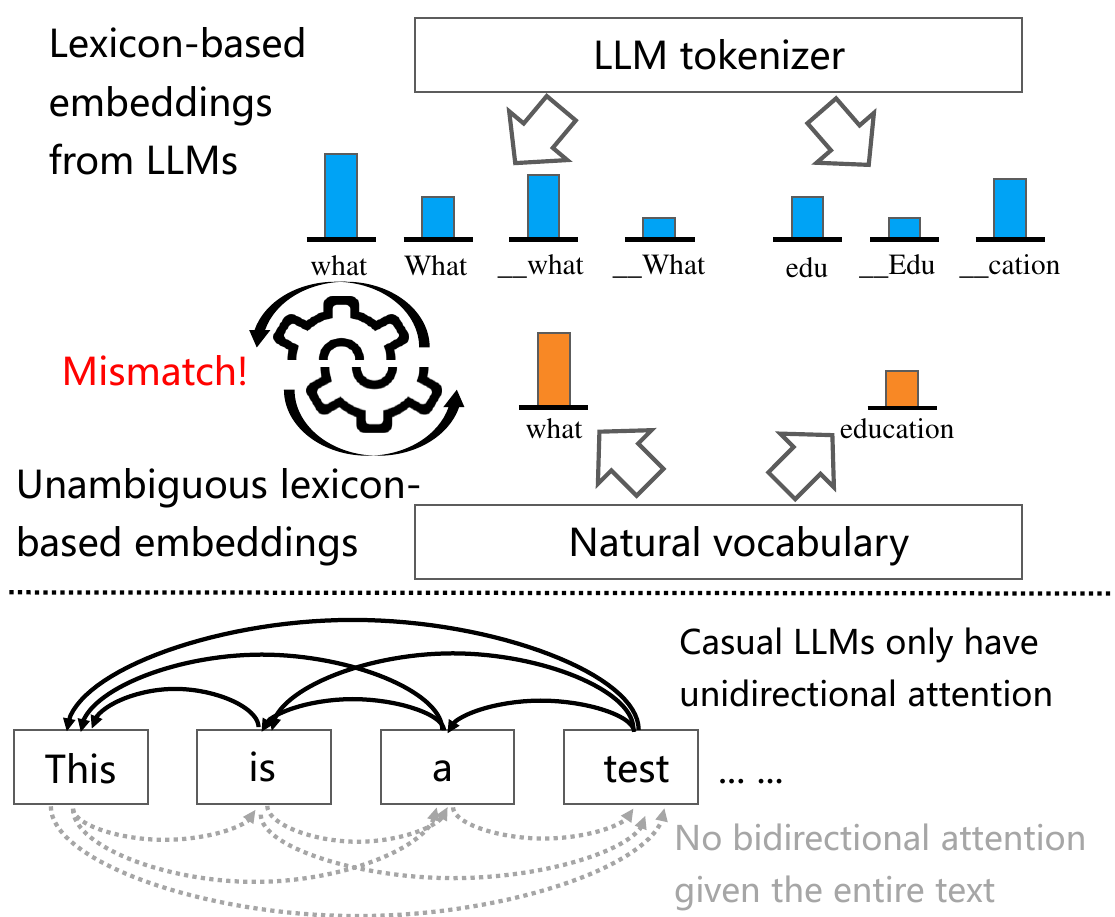}
    \caption{The redundancy and noise in LLM tokenizers, as well as the absence of bidirectional dependencies in causal LLMs motivate \ours.}
    \label{fig:intro_fig}
\end{figure}

While dense embeddings that encode texts into low-dimensional, real-valued latent semantic spaces dominate recent research, lexicon-based embeddings~\citep{spladev2, splade, lexmae, spladev3}
offer distinctive advantages. These high-dimensional representations, where each dimension corresponds to a specific token of the vocabulary, align more closely with the pre-training objectives of language models due to their shared use of the vocabulary space and the language modeling head~\citep{lexmae}. Recent studies have demonstrated that lexicon-based embeddings can surpass their dense counterparts, utilizing masked language models under specific control~\citep{bench_lexicon}.
Additionally, lexicon-based embeddings can offer better transparency, providing clearer insights into the representations via the weight of each token. Moreover, the combination of dense and lexicon-based embeddings has also been shown to be promising in prior studies, as they effectively complement each other~\cite{proposedconceptualframeworkrepresentational,unifier,gao21complement}.

Despite these benefits, lexicon-based embeddings remain underexplored beyond retrieval tasks. To effectively apply them to other tasks, it is essential to address the challenges posed by LLMs, as shown in Fig.~\ref{fig:intro_fig}. The first issue is the inherent redundancy of LLM vocabularies. Since most modern tokenizers rely on subword tokenization (e.g., \texttt{``education''} is split into \texttt{``edu''} and \texttt{``cation''}), it fragments the entire vocabulary space~\citep{impactwordsplitting}. More importantly, semantically equivalent tokens can appear in multiple forms in the tokenizer (e.g., \texttt{``what''}, \texttt{``What''}, \texttt{`` what''} and \texttt{``review''}, \texttt{``reviews''}), introducing inconsistencies and difficulties in lexicon matching (e.g., the choice of whether \texttt{``what''} and/or \texttt{`` what''} appears in the lexical representation should be consistent along with their associated weights). Consequently, recent studies indicate that replacing the original tokenization of BM25~\citep{bm25} with the XLM-R tokenizer~\citep{xlmr} can lead to a significant performance drop due to the noisier vocabulary~\citep{m3}. The second challenge is that LLMs typically employ unidirectional attention during pre-training, where tokens only attend to preceding tokens. This limitation prevents each token from fully leveraging the surrounding context, which is crucial as lexicon-based embeddings are derived from the outputs of all tokens.

To address these challenges, we first explore the potential of LLMs generating embeddings where each dimension corresponds to a token cluster instead of the traditional single token, with each cluster grouping tokens that share similar meanings or stem from the same lexeme. To achieve this, we utilize a simple yet effective approach that directly clusters the token embeddings and leverages the centroids of these clusters as the new token embeddings for the language modeling head.
As shown in Table~\ref{tab:cluster_examples}, the resulting clusters naturally group tokens with similar meanings, forming more coherent and compact embeddings. These cluster-based embeddings can achieve the equivalent feature size as dense embeddings from the same LLM (e.g., 4,000d), which is much smaller than previous lexicon-based embeddings. Such a property i) facilitates the integration of \ours into existing dense frameworks like FAISS, reducing the need for sparsity constraints, and also ii) eliminates computational overhead in tasks such as clustering and classification, where inverted indices are not a natural fit.

Given recent studies highlighting the significant impact of attention mechanisms and pooling strategies on dense embeddings~\citep{bge_en_icl, grit_lm, llm2vec, nv_embed}, we incorporate variants of these two factors in our framework to examine how they affect lexicon-based embeddings. Contrary to prior findings~\citep{bge_en_icl}, which suggest that preserving the original architecture of LLMs typically yields optimal performance for dense embeddings, our results indicate that bidirectional attention is critical for achieving superior performance with lexicon-based embeddings.

Built on these techniques, we introduce \ours, a framework designed to generate low-dimensional lexicon-based embeddings that achieve impressive results across a variety of tasks.
Specifically, our experiments demonstrate that \ours outperforms dense embeddings on the Massive Text Embedding Benchmark (MTEB)~\citep{mteb}, achieving state-of-the-art (SOTA) zero-shot performance among models trained exclusively on public data, as of December 1, 2024.
These results demonstrate that sparse lexicon-based embeddings can perform competitively with dense embeddings across a diverse range of tasks.
Qualitative examples also illustrate that \ours produces grounded and meaningful representations. Our analysis also shows that \ours, even when using 2000 clusters, still outperforms embeddings that leverage the original vocabulary space.  
Efficiency can be further improved through Top-K pruning: by retaining only the top 256 of 4000 dimensions at inference time, \ours maintains high performance with minimal degradation. This sparsification requires no specialized training (e.g., Matryoshka Representation Learning~\cite{mrl}) and arises naturally from its lexicon-based design.
Finally, combining \ours with dense embeddings achieves SOTA performance on the retrieval subset of MTEB (specifically, BEIR).

\section{Related Work}

\paragraph{Lexicon-Based Embeddings.}
Lexicon-based embeddings assign each dimension of the embedding vector to a specific token in the vocabulary. With advances in masked language models, recent studies have demonstrated that lexicon-based embeddings~\citep{mallia2021learning, lin2021briefnotes, zhuang2024fast, spladev2, splade, lexmae, unifiedlexicon, spladev3, thong2023adapting, nguyen-etal-2024-dyvo, lupart2025discollmknowledgedistillation} can deliver strong performance. Among these approaches, SPLADE~\citep{splade, spladev2, spladev3} stands out as one of the most effective methods, even outperforming comparable dense embeddings~\cite{bench_lexicon}. Moreover, lexicon-based embeddings have been shown to complement dense embeddings, with their combination yielding substantial performance improvements~\cite{m3, unifier, proposedconceptualframeworkrepresentational,gao21complement}. Despite these advances, research on lexicon-based embeddings has largely focused on retrieval tasks, leaving other applications relatively such as clustering and classification underexplored.

\paragraph{LLM-Based Embeddings.}
As decoder-only LLMs continue to advance, recent work has investigated their potential for generating dense text embeddings capable of performing well across different tasks.
To align LLMs with text embedding tasks, LLM2Vec~\cite{llm2vec} enables bi-directional attention with masked next-token prediction training and unsupervised contrastive learning, while LLaRA~\cite{llara} leverages an auto-encoding objective to enhance embedding quality. Recent efforts, such as E5-Mistral~\citep{e5_mistral} and Gecko~\citep{gecko}, focus on improving embedding models by using LLMs to generate diverse training data. 
Additionally, GRIT~\cite{grit_lm} explores the combination of contrastive learning and language modeling objectives to train a single LLM that performs well on both embedding and generation tasks.
Meanwhile, studies~\citep{grit_lm, nv_embed, llm2vec, bge_en_icl} highlight the significant influence of architectural choices on embedding model performance, with findings~\citep{bge_en_icl} indicating that retaining the original unidirectional attention often yields the best results.

Research on leveraging LLMs for lexicon-based embeddings remains limited. PromptReps~\citep{promptreps} uses prompt engineering to generate lexicon-based embeddings from LLMs, but performs substantially worse than dense counterparts and approaches that involve fine-tuning like  Mistral-SPLADE~\cite{mistralsplade}.
Furthermore, these efforts are limited to retrieval tasks. We explore how general-purpose lexicon-based embeddings can be produced.

\section{Methodology}
In this section, we first introduce preliminaries for a better understanding of the design of our framework, then formally describe the details of \ours.

\subsection{Preliminaries}

\subsubsection{Lexicon-Based Embeddings Using Masked Language Models}
SPLADE~\cite{splade, spladev2, spladev3} is a representative method that utilizes Masked Language Models (MLMs) and regards the logits from the masked language modeling head as lexicon-based embeddings, leveraging the bidirectional attention.
The MLM produces a sequence of logits \( L = (l_1, l_2, \dots, l_n), l_i \in \mathbb{R}^{|V|} \) given the input sequence, where \( |V| \) is the vocabulary size. Each logit value $l_{ij}$ represents the likelihood of the vocabulary token $j$ being relevant to the position $i$. Specifically, these scores are produced by the language modeling head, which maps the output hidden states to the vocabulary space using the token embedding matrix.

To obtain the lexicon-based embeddings, SPLADE first applies a log-saturation transformation to the logits to scale the weight and enforce it as non-negative,
\begin{equation}
\label{eq_log}
w_{ij} = \log\left(1 + {ReLU}(l_{ij})\right).
\end{equation}
Then it performs max-pooling across logits of all tokens to derive the final weight for each vocabulary token,
\begin{equation}
\label{eq_max_pooling}
w_j = \max_{i \in n} w_{ij}.
\end{equation}

Despite its proven effectiveness, former research on lexicon-based embeddings using MLMs primarily focused on small-scale models, leaving the performance of larger models mostly unexplored.

\subsubsection{Lexicon-Based Embeddings Using Causal Language Models}
Motivated by the growing capability of larger-scaled models, recent works have begun to use causal language models with significantly more parameters, such as LLaMA~\citep{llama} and Mistral~\citep{mistral_7b}, to derive lexicon-based embeddings. Two notable methods are \textbf{PromptReps}~\citep{promptreps} and \textbf{Mistral-SPLADE}~\citep{mistralsplade}, which employ prompts to alleviate the limitations brought by the unidirectional attention.

\paragraph{PromptReps} enables LLMs to generate both dense and lexicon-based embeddings through carefully designed prompts such as \textit{``This sentence [INPUT] means in one word:''}. Dense embeddings are derived from the hidden states of the final token \textit{"}, and lexicon-based ones are the logits for the next token prediction. Nevertheless, such a method relying solely on prompt causes a substantial performance drop of lexicon-based embeddings compared to their dense counterparts, e.g., MRR@10 of 34.15 vs. 41.86 on the MS MARCO dataset. 

\paragraph{Mistral-SPLADE} adapts SPLADE to large causal models like Mistral by using echo prompting~\citep{repetitionimproveslanguagemodel}. It enables full-context visibility of each token by duplicating the input sequence and regards the representations of the second occurrence as the output. Despite getting advancements on the BEIR benchmark, it still lags behind dense embeddings like E5-Mistral~\citep{e5_mistral} and LLM2Vec~\cite{llm2vec}, demonstrating that lexicon-based embeddings using large models cannot solely rely on prompting.

Hence, \ours systematically investigate the architecture of LLMs, including attention mechanisms and pooling methods, rather than exterior prompting. We investigate lexicon-based embeddings from LLMs, not only on retrieval tasks that have been widely examined before, but also on clustering and classification tasks which remain unexplored.

\subsubsection{Tokenization in LLMs}
LLM tokenizers, though designed to cover all possible text forms for the language modeling objective, may hinder the effectiveness of lexicon-based embedding. i) High redundancy can be introduced under the same lexeme and further affect the token matching. E.g., \texttt{``What''}, \texttt{``what''}, and \texttt{`` what''} can be regarded as distinct tokens due to differences in case or whitespace, even though they represent the same word, and may have different weights in the embedding.
ii) Subword fragmentation~\citep{impactwordsplitting} split a common word into pieces like \texttt{``education''} into \texttt{``edu''} and \texttt{``cation''}, posing additional matching complexity. iii) Tokenizers trained on large corpora often include rare tokens, which inflate the vocabulary size and make the embedding larger and slower to match. 

Therefore, instead of directly using the original language modeling head, we simply cluster original tokens to form clusters and use their centroid embeddings to replace the original token embeddings of the language modeling head. This approach reduces the redundancy by merging related tokens and decreases the size of embeddings by using a smaller clustered vocabulary.

\subsection{Framework of \ours}

\begin{figure}[!t]
    \centering
    \setlength{\abovecaptionskip}{0.3cm}
    \setlength{\belowcaptionskip}{-0.3cm}
    \includegraphics[width=0.95\linewidth]{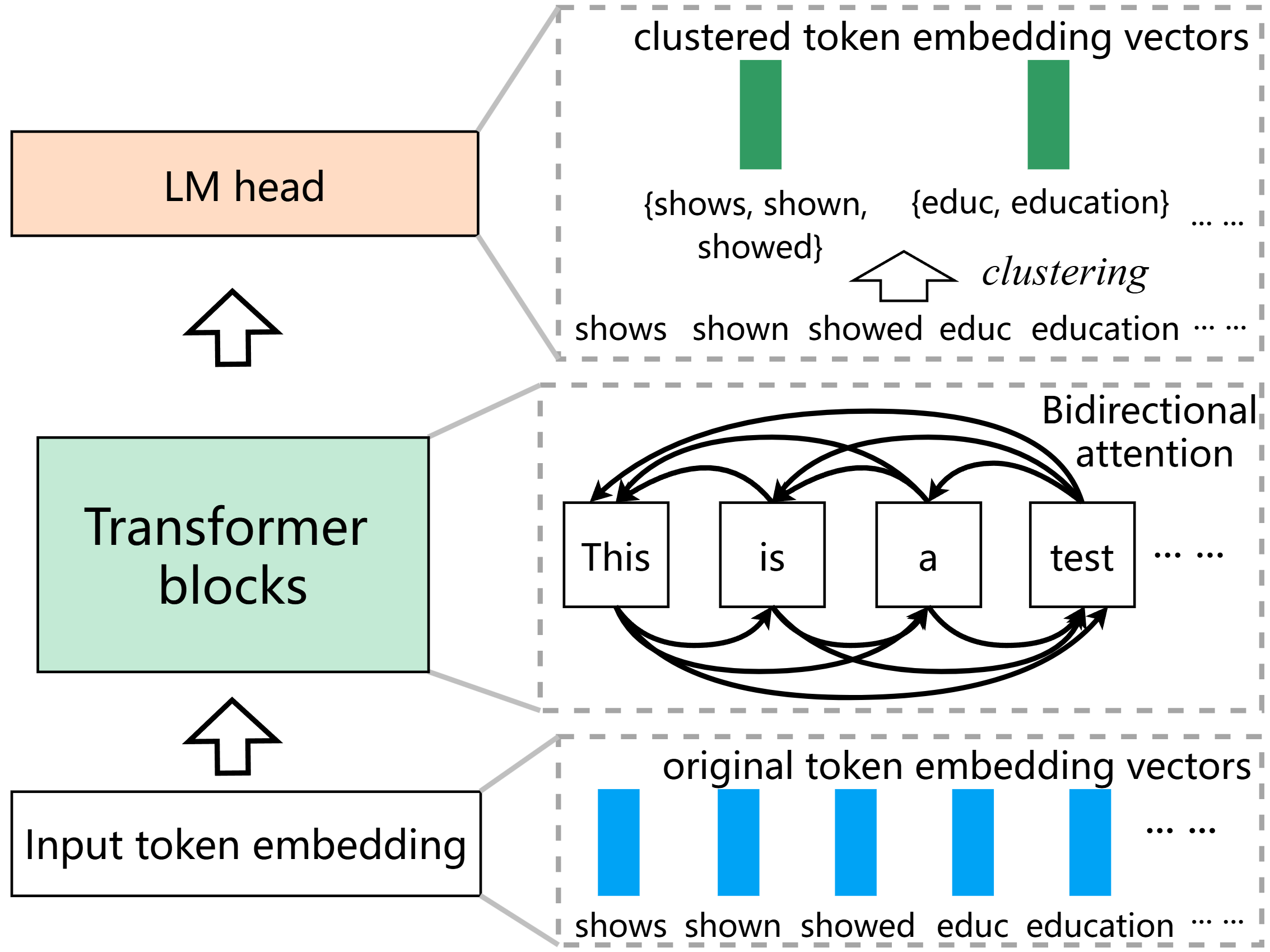}
    \caption{The model framework of \ours. }
    \label{fig:framework}
\end{figure}

After discussing the background, we introduce the framework of our method, as shown in Fig.~\ref{fig:framework}. 

\subsubsection{Architecture Design}\label{method_framework}
\paragraph{Language Modeling Head.}
Motivated by the redundancy and noise in LLM tokenizer mentioned above, \ours assigns weights to groups of tokens with similar meanings, whose effectiveness has been verified in~\citet{tomato}. 
Specifically, we apply K-means clustering~\cite{kmeans} to the token embeddings from the language modeling head, where $k$ is our desired lexicon-based embedding size. Then the original token embeddings in the LM head are replaced by the cluster centroids, while the input token embeddings remain unchanged. This substitution reduces the dimensionality of the lexicon-based embeddings, as the logits now represent scores over clusters rather than the large original vocabulary. See Table~\ref{tab:cluster_examples} and Appendix~\ref{sec:cluster_results} for detailed cluster results.

\paragraph{Attention Mechanism.}
Given the former illustration of the limitations of unidirectional attention in typical causal LLMs, we emphasize it restricts the visibility of each token to the entire context. Hence, unlike previous works that rely on non-fundamental solutions like prompt engineering, we address this issue by directly modifying attention to be bidirectional during fine-tuning, which makes prompt design easier and inference more efficient.

\subsubsection{Representation Generation}
Following~\citet{e5_mistral} and~\citet{bge_en_icl}, given a raw query-passage pair $(q, p)$ for a specific embedding task, we first construct the instructed query input text as 
\begin{equation}
q_\mathrm{ins} = \langle \mathrm{Instruct} \rangle \{\mathrm{task\_definition}\} \langle \mathrm{query} \rangle \{q\}.
\end{equation}
Here \textit{task\_definition} refers to the definition of the specific embedding task, guiding the model to adapt towards that task. On the other hand, the input of the passage part is solely the original text. Following~\citet{e5_mistral} and~\citet{bge_en_icl}, a [EOS] token is also appended to the end of the sequence.

We then feed this input into the modified LLM, and derive a series of logits vectors \( L = (l_1, l_2, \dots, l_n), l_i \in \mathbb{R}^k \), where $n$ is the sequence length and $k$ is our clustering size. To obtain the final embeddings, following~\citet{spladev2}, log-saturation and max-pooling are applied to $L$ along the sequence dimension as in Eq.~\ref{eq_log}, and ~\ref{eq_max_pooling}. 

We only employ tokens corresponding to the original query $q$ to derive the output of the query, avoiding the noise brought by \textit{task\_definition} tokens, inspired by~\citet{llm2vec}. Moreover, considering the autoregressive nature of LLMs, where each logit is used for the prediction of the subsequent position, we shift the logits during pooling. Specifically, we regard the logit corresponding to the neighbor on the left of each token as its feature during computation.

\subsubsection{Training}
Recent research has explored various complex methods for training embedding models. 
For example, NV-Embed-v2~\cite{nv_embed} employs a two-stage training pipeline while also incorporating positive-aware hard-negative mining and synthetic data generation. In contrast, for simplicity and fair comparison, the training of \ours strictly adheres to the training procedure of BGE-en-ICL~\cite{bge_en_icl}, a SOTA LLM-based dense embedding model. It uses a single-stage training process and relies exclusively on publicly available data.

Given a processed input pair ($q_\mathrm{ins}$, $p$), we utilize the InfoNCE loss as our objective,
\begin{equation}
\mathcal{L} = -\log \frac{\exp(\mathrm{sim}(q_\mathrm{ins}, p) / \tau)}{\exp(\frac{\mathrm{sim}(q_\mathrm{ins}, p)}{\tau}) + \sum_{j=1}^{N} \exp(\frac{\mathrm{sim}(q_\mathrm{ins}, p^-_j)}{\tau})}
\end{equation}
Here $p^-_j$ and $N$ denote the negative passage and the number of negative passages, respectively. $\mathrm{sim}()$ is the cosine similarity function, defined as $\mathrm{sim}() = \mathrm{cos}(h_{q_\mathrm{ins}}, h_p)$, where $h_{q_\mathrm{ins}} \in \mathbb{R}^k$ and $h_p \in \mathbb{R}^k$ are the lexicon-based embeddings from the LLM for the instructed query and passage. The temperature $\tau$ is set to 0.02 in our experiments.

\begin{table*}[t]
\centering
\small
\setlength{\abovecaptionskip}{0.2cm}
\setlength{\belowcaptionskip}{-0.1cm}
\setlength{\tabcolsep}{5pt}
\setlength{\extrarowheight}{0.5pt}
\begin{tabular}{ll|ccccccc|c}
\toprule
Task & \#Dims & Retr. & Rerank. & Clust. & PairClass. & Class. & STS & Summ. & Avg. \\
\# of datasets $\rightarrow$ & & 15 & 4 & 11 & 3 & 12 & 10 & 1 & 56 \\ \hline 
\multicolumn{10}{c}{Non-Fully Public Training Data} \\ \hline
E5-mistral-7b-instruct & 4096 & 56.90 & 60.21 & 50.26 & 88.34 & 78.47 & 84.66 & 31.40 & 66.63 \\
Linq-Embed-Mistral & 4096 & 60.19 & 60.29 & 51.42 & 88.35 & 80.20 & 84.97 & 30.98 & 68.17 \\
voyage-large-2-instruct & 1024 & 58.28 & 60.09 & 53.35 & 89.24 & 81.49 & 84.31 & 30.84 & 68.23  \\
stella\_en\_400M\_v5 & 8192 & 58.97 & 60.16 & 56.70 & 87.74 & 86.67 & 84.22 & 31.66 & 70.11 \\
gte-Qwen2-7B-instruct & 3584 & 60.25 & 61.42 & 56.92 & 85.79 & 86.58 & 83.04 & 31.35 & 70.24 \\
SFR-Embedding-2\_R & 4096 & 60.18 & 60.14 & 56.17 & 88.07 & 89.05 & 81.26 & 30.71 & 70.31 \\
stella\_en\_1.5B\_v5 & 8192 & 61.01 & 61.21 & 57.69 & 88.07 & 87.63 & 84.51 & 31.49 & 71.19 \\
NV-Embed-v2 & 4096 & 62.65 & 60.65 & 58.46 & 88.67 & 90.37 & 84.31 & 30.70 & 72.31 \\ \hline
\multicolumn{10}{c}{Fully Public Training Data} \\ \hline
LLM2Vec-Mistral-supervised & 4096 & 55.99 & 58.42 & 45.54 & \textbf{87.99} & 76.63 & 84.09 & 29.96 &  64.80  \\
GritLM-7B & 4096 & 57.41 & 60.49 & 50.61 & 87.16 & 79.46 & 83.35 & 30.37 &  66.76  \\
NV-Embed-v1 & 4096 & 59.36 & 60.59 & 52.80 & 86.91 & 87.35 & 82.84 & \textbf{31.20} & 69.32  \\
bge-multilingual-gemma2 & 3584 & 59.24 & 59.72 & 54.65 & 85.84 & 88.08 & 83.88 & \textbf{31.20} & 69.88 \\
{BGE-en-ICL (zero-shot)} & 4096 & \underline{61.67} & 59.66 & 57.51 & 86.93 & \textbf{88.62} & 83.74 & 30.75 & \underline{71.24} \\
\ours-4000 (\textbf{\textit{Ours}}) & 4000 & 60.76 & \underline{60.86} & \underline{57.92} & 87.93 & 88.13 & \underline{84.35} & 31.56 & 71.22 \\
\ours-8000 (\textbf{\textit{Ours}}) & 8000 & \textbf{61.86} & \textbf{60.91} & \textbf{58.02} & \underline{87.98} & \underline{88.43} & \textbf{84.67} & 29.54 & \textbf{71.63} \\
\bottomrule
\end{tabular}
\caption{Top-performing models on the MTEB leaderboard as of December 1, 2024 compared to \ours. \#Dims refers to the embedding dimensions. Abbreviations: Retr. = Retrieval; Rerank. = Reranking; Clust. = Clustering; PairClass. = Pair Classification; Class. = Classification; STS = Semantic Textual Similarity; Summ. = Summarization. The best and the second best results using public data are in \textbf{bold} and \underline{underlined} font, respectively.}
\label{table:mteb}
\end{table*}

\section{Experiments}
\subsection{Setup}
To ensure a fair comparison between dense embeddings and \ours, we strictly adhere to the training recipe of the SOTA dense model, BGE-en-ICL.

\begin{table*}[ht!]
\centering
\begin{small}
\setlength{\tabcolsep}{4.5pt}
\setlength{\extrarowheight}{0.5pt}
\setlength{\abovecaptionskip}{-0cm}
\setlength{\belowcaptionskip}{-0.3cm}
\begin{tabular}{ll|cccccccc|c}
\toprule
Domain & \#Dims & wiki & web & news & healthcare & law & finance & arxiv & msmarco & Avg. \\
\# of datasets $\rightarrow$ & & 1 & 1 & 1 & 1 & 1 & 1 & 1 & 1 & 8 \\ \hline
E5-mistral-7b-instruct & 4096 & 61.67 & 44.41 & 48.18 & 56.32 & 19.32 & 54.79 & 44.78 & 59.03 & 48.56 \\
Linq-Embed-Mistral & 4096 & 61.04 & 48.41 & 49.44 & \textbf{60.18} & 20.34 & 50.04 & 47.56 & 60.50 & 49.69 \\
NV-Embed-v1 & 4096 & 62.84 & 50.42 & 51.46 & 58.53 & 20.65 & 49.89 & 46.10 & 60.27 & 50.02 \\
gte-Qwen2-7B-instruct & 3584 & 63.46 & 51.20 & 54.07 & 54.20 & 22.31 & \textbf{58.20} & 40.27 & 58.39 & 50.26 \\
stella\_en\_1.5B\_v5 & 8192 & 61.99 & 50.88 & 53.87 & 58.81 & 23.22 & \underline{57.26} & 44.81 & 61.38 & 51.53 \\
SFR-Embedding-Mistral & 4096 & 63.46 & 51.27 & 52.21 & 58.76 & 23.27 & 56.94 & 47.75 & 58.99 & 51.58 \\
NV-Embed-v2 & 4096 & \underline{65.19} & 52.58 & 53.13 & \underline{59.56} & 25.00 & 53.04 & \textbf{48.94} & 60.80 & 52.28 \\
BGE-en-ICL (zero-shot) & 4096 & 64.61 & \underline{54.40} & \underline{55.11} & 57.25 & \underline{25.10} & 54.81 & \underline{48.46} & \textbf{63.71} & \textbf{52.93} \\
\ours-4000 (\textbf{\textit{Ours}}) & 4000 & 62.60 & 52.06 & 52.49 & 57.23 & 24.08 & 48.87 & 43.78 & 61.17 & 50.28 \\
\ours-8000 (\textbf{\textit{Ours}}) & 8000 & \textbf{65.50} & \textbf{54.52} & \textbf{55.16} & 58.20 & \textbf{25.62} & 54.57 & 45.45 & \underline{63.00} & \underline{52.75} \\
\bottomrule
\end{tabular}
\end{small}
\caption{QA performance on AIR-Bench 24.04 (English) across different models, where nDCG@10 is used as the metric. \#Dims refers to the embedding dimensions. The best and the second best results across all models are in \textbf{bold} and \underline{underlined} font, respectively.}
\label{table:air_qa}
\end{table*}

\begin{table*}[t]
\centering
\setlength{\tabcolsep}{1pt}
\setlength{\abovecaptionskip}{0.1cm}
\setlength{\belowcaptionskip}{-0.3cm}
\resizebox{\linewidth}{!}
{%
\begin{tabular}{ll}
\toprule
{Text} & {Top-weighted clusters}\\
\midrule
{most dependable affordable cars} & {(\texttt{cars}, \texttt{Cars}), (\texttt{cheap}, \texttt{affordable}), (\texttt{reliable}, \texttt{reli}), (\texttt{depend}, \texttt{depends), (\texttt{aff}, \texttt{afford})}}\\
{fastest growing bonsai trees} & (\texttt{faster}, \texttt{fastest}), (\texttt{grow}, \texttt{growing}), (\texttt{fast}, \texttt{Fast}), (\texttt{tree}, \texttt{trees}), (\texttt{quickly}, \texttt{rapid}) \\
{causes of hypoxia in adults} & (\texttt{adult}, \texttt{adults}), (\texttt{oxygen}, \texttt{oxy}), (\texttt{cause}, \texttt{caused}), (\texttt{hyp}, \texttt{yp}), (\texttt{ox}, \texttt{Ox}) \\
{weather in lisbon april} & (\texttt{Portug}, \texttt{Portuguese}), (\texttt{bon}, \texttt{Bon}), (\texttt{weather}, \texttt{rather}), (\texttt{Spring}, \texttt{spring}), (\texttt{AP},  \texttt{\#AP}) \\
{other hot flashes causes} & (\texttt{hot}, \texttt{Hot}), (\texttt{cause}, \texttt{causes}), (\texttt{flash}, \texttt{Flash}), (\texttt{flush}, \texttt{\#flush}), (\texttt{heat}, \texttt{Heat}) \\
\bottomrule
\end{tabular}}
\caption{Qualitive examples of \ours-8000. For each example, the top-5 clusters with the largest weights in the embeddings are shown, with two tokens from each cluster included. \# denotes a whitespace character.} \label{tab:qualitative_examples}
\end{table*}

\paragraph{Model Setup.}
The Mistral-7B-v0.1~\cite{mistral_7b} model is used as the backbone in \ours, in line with recent works such as BGE-en-ICL~\cite{bge_en_icl}, E5-Mistral~\cite{e5_mistral}, NV-Embed-v2~\cite{nv_embed}, and LLM2Vec~\cite{llm2vec}. To investigate the effect of different clustering sizes to consolidate the output token embeddings, we set $k$ in KMeans clustering to 4,000 and 8,000 clusters, referred to as \ours-4000 and \ours-8000, respectively. \ours-4000 can output 4000-d embeddings, which is comparable to the 4096-d dense embeddings produced by the same backbone LLM.

\paragraph{Training Data.}
We directly utilize the publicly available training data provided by BGE-en-ICL. This dataset is a mixture of retrieval, reranking, clustering, classification, and semantic textual similarity (STS) tasks. Details about the training data can be found in Appendix~\ref{data_details}. We use the same set of task instructions as BGE-en-ICL, refer to Appendix~\ref{sec:instructions} for details.

\paragraph{Training Configurations.}
Following BGE-en-ICL, our model is trained for one epoch using LoRA~\cite{lora}, where the LoRA rank is 32 and the alpha is 64, and the learning rate is set to 1e-4. Each training sample is composed of 1 positive and 7 hard negatives. For retrieval tasks, we use a batch size of 512, whereas a batch size of 256 is used for the rest tasks. All data are drawn from the same dataset within the same batch. In retrieval tasks, we employ in-batch negatives and apply a KL-divergence loss to distill ranking scores from the BGE-reranker model. The maximum length for both the query and passage is set to 512. We deviate from BGE-en-ICL by omitting in-context learning samples during training and concentrate on zero-shot scenarios solely, in order to evaluate performance without extraneous signals.

\paragraph{Evaluations.}
We evaluate the performance of various embedding models using MTEB~\cite{mteb} and AIR-Bench~\cite{air_bench}. MTEB is a comprehensive text embedding benchmark encompassing seven task types across a total of 56 datasets. AIR-Bench, on the other hand, spans diverse domains for retrieval tasks, including law, healthcare, and books, having no overlap with MTEB. Notably, the ground truth for the test set in AIR-Bench is hidden, and we use the 24.04 version to assess the model's out-of-domain capabilities. 

We compare \ours to numerous baselines, including E5-mistral-7b-instruct~\cite{e5_mistral}, NV-Embed-v1/v2~\cite{nv_embed},
gte-Qwen2-7B-instruct~\cite{gte_qwen},
LLM2Vec~\cite{llm2vec}, SFR-Embedding-2\_R~\cite{SFR-embedding-2}, 
GritLM-7B~\cite{grit_lm}, and BGE-en-ICL~\cite{bge_en_icl}. The results of PromptReps and Mistral-Splade are excluded, as they are designed specifically for retrieval tasks and their performance falls below of the weakest baseline, namely LLM2Vec-Mistral-supervised. Some of the baselines use private data during training or involve in-context learning. To ensure a fair comparison, we focus on \textbf{zero-shot scenarios} where no few-shot sample is included in the prompt, e.g., BGE-en-ICL.

\subsection{Main results}
\paragraph{MTEB.}
Table~\ref{table:mteb} demonstrates the results of a variety of models on MTEB. \ours-8000 achieves the highest average performance among all models trained on fully public data. Notably, \ours-8000 outperforms BGE-en-ICL, its dense embedding counterpart trained with the same data and hyperparameters, with consistently higher performance in 6 out of 7 categories of tasks. \ours-4000 yields comparable performance to BGE-en-ICL and both produce representations of similar sizes, but our lexicon-based method can deliver better transparency. Furthermore, \ours-8000 ranks second among all models in overall average performance. The leading model, NV-Embed-v2, attains high performance through a significantly more complex training pipeline, which includes a two-stage training pipeline, positive-aware hard-negative mining, and synthetic data generation. By contrast, \ours uses fully public data and adopts a simpler training procedure following BGE-en-ICL.

\begin{table}[t]
\centering
\small
\setlength{\tabcolsep}{0.5pt}
\setlength{\abovecaptionskip}{0.1cm}
\setlength{\belowcaptionskip}{-0.3cm}
\resizebox{\linewidth}{!}
{%
\begin{tabular}{l}
\toprule
{Clusters}\\
\midrule
{\texttt{quickly}, \texttt{rapid}, \texttt{rapidly}, \texttt{swift}}  \\ \hline
{\texttt{cannot}, \texttt{impossible}, \texttt{Unable}, \texttt{Cannot}}, \texttt{Unable} \\ \hline
{\texttt{shows}, \texttt{shown}, \texttt{showed}, \texttt{showing}}   \\ \hline
{\texttt{review}, \texttt{Review}, \texttt{reviews}, \texttt{reviewed}, \texttt{Reviews} }  \\ \hline
{\texttt{educ}, \texttt{education}, \texttt{Educ}, \texttt{Education}, \texttt{educational}}, \texttt{Edu}  \\
\bottomrule
\end{tabular}}
\caption{
Cluster examples of \ours-8000. Each row presents tokens belonging to a single cluster. More cluster examples are provided in Appendix~\ref{sec:cluster_results}.}\label{tab:cluster_examples}
\end{table}

\paragraph{AIR-Bench.}
We also evaluate \ours along with baselines on the QA tasks of AIR-Bench. As shown in Table~\ref{table:air_qa}, \ours-8000 outperforms the top-performing model on MTEB, NV-Embed-v2, demonstrating its promising generalization capabilities. Despite slightly lagging behind its dense counterpart BGE-en-ICL, \ours still remains competitive in several sub-tasks. However, \ours-4000 performs less competitively, potentially because a smaller number of clusters may result in over-generalized clusters and information loss.

\subsection{Qualitative Examples}
We present some clustering results in Table~\ref{tab:cluster_examples}. These examples illustrate that: i) \ours groups semantically equivalent tokens (e.g., ``\texttt{rapid}'' and ``\texttt{quickly}'', ``\texttt{cannot}'' and ``\texttt{impossible}''); ii) it groups morphologically similar tokens (e.g., ``\texttt{shows}'' and ``\texttt{showed}''); and iii) it groups uppercase/lowercase variants and whole-word/subword forms (e.g., ``\texttt{review}'' and ``\texttt{Review}'', ``\texttt{Edu}'' and ``\texttt{education}''). Such an observation highlights the effectiveness of clustering in reducing the redundancy and noise of the tokenizer.

In addition, qualitative examples from MS MARCO of \ours-8000 embeddings are given in Table~\ref{tab:qualitative_examples}. The top-5 clusters with the largest weights in the embeddings are presented for each sample, where two tokens from each cluster are included. Obviously, these clusters are highly semantically relevant to the input texts, which can be regarded as some keywords. There are also interesting findings that the embeddings show some deep understanding of the text such as ``\texttt{oxygen}'' in response to the input ``\texttt{causes of hypoxia in adults}'', and some knowledge expansion capabilities like ``\texttt{Portuguese}'' and ``\texttt{spring}'' for the input ``\texttt{weather in lisbon april}''. These qualitative samples demonstrate that lexicon-based embeddings from LLMs capture more than shallow token meanings.

\begin{table*}[t]
\begin{center}
\begin{footnotesize}
\setlength{\tabcolsep}{5pt}
\setlength{\abovecaptionskip}{-0.2cm}
\setlength{\belowcaptionskip}{-0.3cm}
\begin{tabular}{l|ccccccc|c}
\toprule
Task & Retr. & Rerank. & Clust. & PairClass. & Class. & STS & Summ. & Avg. \\
\# of datasets $\rightarrow$ & 1 & 1 & 1 & 1 & 1 & 1 & 1 & 7 \\ \hline
\multicolumn{9}{c}{Unidirectional Attention} \\ \hline
Last-token pooling & 73.84 & 65.19 & 60.46 & 96.69 & 58.66 & 89.26 & 30.05 & 67.73 \\
Sum-pooling & 72.46 & 59.57 & 50.55 & 89.90 & 54.64 & 80.55 & 29.70 & 62.48 \\
Max-pooling & 75.18 & 59.68 & 50.93 & 92.06 & 57.58 & 82.74 & 30.89 & 64.15 \\ \hline
\multicolumn{9}{c}{Bidirectional Attention} \\ \hline
Last-token pooling & 76.89 & 64.21 & 61.57 & 96.62 & 58.33 & 88.72 & 30.72 &  68.15 \\
Sum-pooling & 75.65 & 63.64 & 61.77 & 96.97 & 60.05 & 89.58 & 30.98 & 68.38  \\
Max-pooling & 76.19 & 64.53 & 63.05 & 97.03 & 62.30 & 88.92 & 31.49 & 69.07 \\
\bottomrule
\end{tabular}
\end{footnotesize}
\end{center}
\caption{Influence of attention mechanisms and pooling methods.}
\vspace{-0.3cm}
\label{table:architecture}
\end{table*}

\section{Analysis}
In this section, we conduct a detailed investigation of \ours design decisions. For the sake of computational resources, we reduce the training data of each dataset to 10\% of its original size and limit both query and passage lengths to a maximum of 128 tokens. Besides, we use the same MTEB subset as \citet{repurposing} for faster evaluation, as it correlates well with the overall performance of MTEB (details in Appendix~\ref{sec:subset_details}).

\begin{figure}[h]
    \centering
    \setlength{\abovecaptionskip}{0cm}
    \setlength{\belowcaptionskip}{-0.3cm}
    \includegraphics[width=1\linewidth]{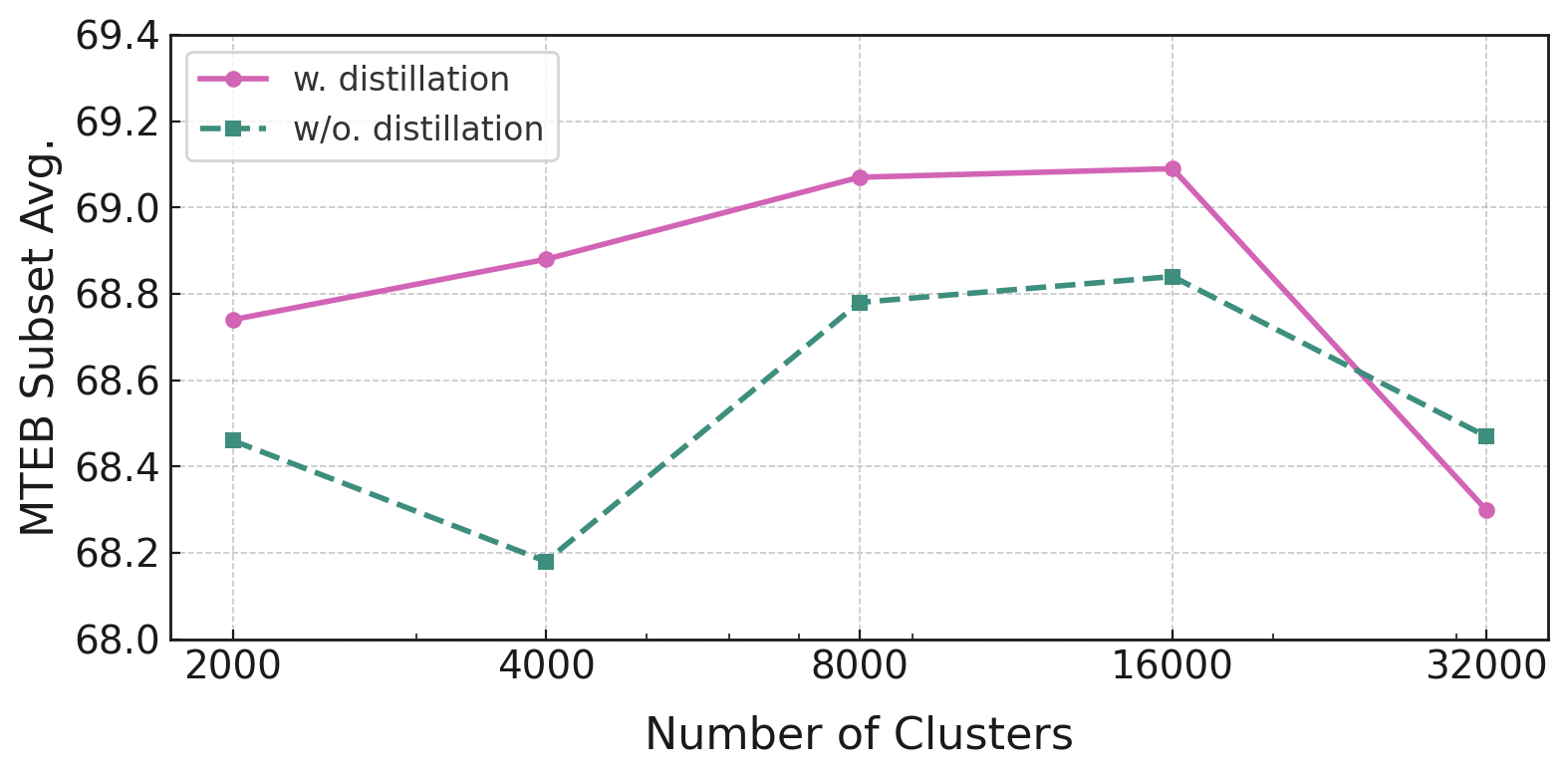}
    \caption{Influence of the number of clusters. The configuration with 32,000 clusters retains the original token embeddings without clustering. \texttt{w/o. distillation} refers to training with only the contrastive loss, excluding the distillation loss.}
    \label{fig:number_of_clusters}
\end{figure}

\subsection{Influence of the Number of Clusters}
We investigate how the number of clusters $k$ affects performance, as shown in Figure~\ref{fig:number_of_clusters}. The configuration with 32,000 clusters corresponds to retaining the original token embeddings without applying clustering. Reducing the number of clusters generally improves performance, with notable gains observed at 8,000 and 16,000 clusters. We also show the performance of models trained without the distillation loss, using only contrastive learning. Across all values of $k$ except 32,000, distillation provides a consistent benefit. A configuration of 8,000 clusters strikes a good balance between effectiveness and efficiency (dimensionality). Consequently, we employ $k=8,000$ in the subsequent experiments.

\subsection{Influence of Model Architecture}
We investigate the effects of the attention mechanism and pooling methods, as illustrated in Table~\ref{table:architecture}. For attention, we examine both unidirectional and bidirectional attention. Regarding pooling strategies, we assess max-pooling, sum-pooling, and last-token pooling. The results highlight the critical role of bidirectional attention in achieving strong performance with lexicon-based embeddings, as evidenced by its superiority across all pooling methods. Among these pooling methods, max-pooling emerges as the most effective strategy. This finding partially explains the poor performance of lexicon-based embeddings from PromptReps, which relies on last-token pooling with unidirectional attention.

\begin{table}[h]
\centering
\footnotesize
\setlength{\tabcolsep}{4pt}
\setlength{\abovecaptionskip}{0.3cm}
\setlength{\belowcaptionskip}{-0.3cm}
\begin{tabular}{l|c|c}
\toprule
Model & \#Active Dims & BEIR Avg. \\
\midrule
BGE-en-ICL & 4096 & 61.67 \\
Gecko & 768 & 55.67 \\
Gecko & 256 & 52.43 \\
\midrule
LENS-4000 & / & 60.76 \\
LENS-4000 & 768 & 60.01 \\
LENS-4000 & 512 & 59.33 \\
LENS-4000 & 256 & 57.19 \\
\bottomrule
\end{tabular}
\caption{
Top-K pruning results on BEIR. LENS-4000 is evaluated by retaining only the top-K activated dimensions at inference time, without retraining. When pruning from 768 to 256 dimensions, LENS exhibits a smaller performance drop than Gecko, despite Gecko is explicitly trained to support multiple embedding sizes.
}
\label{tab:topk_pruning}
\end{table}

\begin{table*}[!t]
\setlength{\abovecaptionskip}{-0cm}
\setlength{\belowcaptionskip}{-0.3cm}
\begin{center}
\begin{footnotesize}
\setlength{\tabcolsep}{1.5pt}
\setlength{\extrarowheight}{1pt}
\begin{tabular}{l|ccccccccccccccc|c}
\hline
Dataset & ARG & CLI & CQA & DBP & FEV & FIQ & HOT & MSM & NFC & NQ & QUO & SCD & SCF & TOU & COV & Avg. \\ \hline
BGE-en-ICL & 82.76 & 45.35 & 47.23 & 50.42 & 91.96 & 58.77 & 84.98 & 46.72 & 40.69 & 73.85 & 91.02 & 25.25 & 78.33 & 29.67 & 78.11 & 61.67 \\
\ours-8000 (\textit{\textbf{Ours}}) & 76.02 & 45.77 & 48.67 & 49.75 & 92.32 & 61.57 & 85.71 & 47.24 & 40.61 & 74.64 & 90.79 & 28.54 & 79.75 & 29.34 & 77.18 & 61.86 \\
NV-Embed-v2 & 70.07 & 45.39 & 50.24 & 53.50 & 93.75 & 65.73 & 85.48 & 45.63 & 45.17 & 73.57 & 89.04 & 21.90 & 80.13 & 31.78 & 88.44 & 62.65 \\
\ours (\textit{\textbf{Ours}}) + BGE & 81.37 &  \textbf{47.14} & {48.57} & \textbf{51.79} & \textbf{93.12} & \textbf{62.00} & \textbf{87.12} & \textbf{47.66} & \textbf{41.55} & \textbf{75.81} & \textbf{91.07} & 28.41 & \textbf{80.19} & \textbf{30.51} & \textbf{78.72} & \textbf{63.00} \\
\hline
\end{tabular}
\end{footnotesize}
\end{center}
\caption{Results in terms of nDCG@10 on the retrieval subset (i.e. BEIR) of MTEB. We use the first three letters of each dataset's name as its abbreviation, except SCIDOCS (abbreviated as SCD) and SciFact (abbreviated as SCF). \textbf{Bolded values} indicate datasets where the combinations outperform both \ours-8000 and BGE-en-ICL individually. On 12 out of 15 datasets, combining \ours-8000 and BGE-en-ICL results in improved performance. 
}
\label{table:combine}
\end{table*}

\subsection{Top-K Pruning and Natural Sparsity}

Table~\ref{tab:topk_pruning} shows the effect of Top-K pruning on LENS embeddings. Each embedding dimension in LENS corresponds to a distinct lexical cluster, resulting in disentangled representations that naturally support sparsification. We apply Top-K pruning at inference time by retaining only the $K$ dimensions with the highest activation values in the embedding, setting the rest to zero. We use LENS-4000 in this analysis due to its relatively low dimensionality, which allows Top-K pruning to be particularly effective. This approach requires no retraining and produces compact, sparse representations.
We compare LENS to Gecko~\cite{gecko}, which is trained with Matryoshka Representation Learning~\cite{mrl} to explicitly support multiple embedding sizes. In contrast, LENS is trained once and pruned post hoc.

\subsection{Hybrid Lexicon-Dense Embeddings}
Previous studies have demonstrated that lexicon-based embeddings and dense embeddings are complementary, and combining them can lead to significant performance improvements. In this section, we explore the effectiveness of combining \ours with BGE-en-ICL, both trained on the same data but representing different types of embeddings. To evaluate general-use cases, we concatenate the two embeddings into a single embedding, without applying any additional operations. We hypothesize that enhanced performance could be achieved by tuning the combination weights of the two embeddings.

The results are presented in Table~\ref{table:combine}. Combining \ours-8000 with BGE-en-ICL yields a substantial performance improvement, increasing from 61.67/61.86 to 63.00, which surpasses NV-Embed-v2 and achieves SOTA results on the retrieval subset of MTEB as of December 1, 2024. Furthermore, such an improvement is consistent, as evidenced by performance gains on 12 out of 15 datasets.

\subsection{Failure Cases of LENS}

\begin{table}[t]
\centering
\setlength{\tabcolsep}{1pt}
\setlength{\abovecaptionskip}{0.1cm}
\setlength{\belowcaptionskip}{-0.3cm}
\resizebox{\linewidth}{!}
{%
\begin{tabular}{ll}
\toprule
{Text} & {Problematic Top-10 Weighted Clusters}\\
\midrule
{how much does an average person make for tutoring} & (\texttt{tuple}, \texttt{Tuple}) \\
{what county is rossville, ga in} & (\texttt{count}, \texttt{Count}), (\texttt{COUNT}), (\texttt{SSL}, \texttt{ssl}) \\
{what causes tomato dry rot in tomatoes} & (\texttt{tom}), (\texttt{Tom}, \texttt{Tommy}), (\texttt{rott}, \texttt{\#rott})  \\
{what is a hangar} & (\texttt{har}, \texttt{\#har}), (\texttt{anger}, \texttt{angers}), (\texttt{tang}) \\
\bottomrule
\end{tabular}}
\caption{Examples of failure cases from \ours-8000, where some of the top-10 weighted dimensions are associated with clusters that are lexically similar to the query terms but semantically unrelated. Each cluster is illustrated with up to two representative tokens.}
\label{tab:failure_examples}
\end{table}

The lexicon-grounded nature of LENS enables inspection of failure cases by examining the most influential token clusters for each input text. Table~\ref{tab:failure_examples} shows several examples from LENS-8000 where the top-weighted dimensions are dominated by clusters that are lexically similar but semantically irrelevant.

Many of these issues stem from the subword tokenization used in LLMs, where words are broken into fragments that may overlap in form but not in meaning. For example, the text ``\texttt{how much does an average person make for tutoring}'' activates the cluster (``\texttt{tuple}'', ``\texttt{Tuple}''), likely due to subword overlap with ``\texttt{tutoring}'' rather than any meaningful connection. Similarly, ``\texttt{what is a hangar}'' triggers clusters such as (``\texttt{har}'', ``\texttt{har}'') and (``\texttt{anger}'', ``\texttt{angers}''), which are unrelated in meaning but happen to resemble parts of the input string.

To address these limitations, a promising direction is to shift from token-level to entity-level representations~\cite{nguyen-etal-2024-dyvo}, associating embedding dimensions with semantically meaningful whole words or phrases rather than isolated subword fragments. Another complementary direction is to explicitly down-weight or drop high-frequency tokens that tend to appear in many clusters across the corpus, as these often reflect common morphemes or generic terms that contribute little to semantic discrimination but introduce noise into the learned dimensions.

\section{Conclusion}
In this work, we introduce \ours, a simple yet effective framework for generating lexicon-based text embeddings using LLMs. Our approach leverages token embedding clustering to address the redundancy challenges inherent in LLM tokenizers, while also enabling bidirectional attention to fully unlock the potential of LLMs. Extensive experiments demonstrate the promising effectiveness and generalization capabilities of \ours compared to SOTA dense embeddings. Qualitative examples illustrate that \ours produces embeddings that are grounded and demonstrate a deep understanding of the input. Further analyses show the superiority of fusing lexicon-based \ours and dense embeddings, which surpasses each individual model on the retrieval subset of MTEB (i.e., BEIR). 

\section*{Limitations}
We acknowledge the following limitations of our work. First, our training and evaluation are limited to English, leaving multilingual datasets, such as Miracl~\citep{zhang2023miracl}, unexplored. This restricts the generalizability of our findings to non-English contexts. Second, we applied \ours exclusively to the widely used Mistral-7B model, leaving other models unexplored. Additionally, compared to previous lexicon-based models like SPLADE, utilizing LLMs as the backbone significantly increases computational costs.

\section*{Acknowledgements}
This research was supported by the Hybrid Intelligence Center, a 10-year program funded by the Dutch Ministry of Education, Culture and Science through the Netherlands Organisation for Scientific Research, \url{https://hybrid-intelligence-centre.nl}, and project VI.Vidi.223.166 of the NWO Talent Programme which is (partly) financed by the Dutch Research Council (NWO).

\bibliography{custom}

\appendix
\newpage
\section{Appendix}
\label{sec:appendix}
\subsection{Clustering results}\label{sec:cluster_results}
\begin{table}[h]
\centering
\small
\setlength{\tabcolsep}{0.5pt}
\setlength{\belowcaptionskip}{-0.3cm}
\resizebox{\linewidth}{!}
{%
\begin{tabular}{l}
\toprule
\textbf{Clusters}\\
\midrule
{\texttt{impact}, \texttt{Impact}, \texttt{impacts}}  \\ \hline
{\texttt{Entity}, \texttt{entity}, \texttt{\#Entity}, \texttt{Entities}}, \texttt{entities} \\ \hline
{\texttt{TV}, \texttt{television}, \texttt{tv}, \texttt{Television}, \texttt{televis}}   \\ \hline
{\texttt{comfort}, \texttt{comfortable}, \texttt{\#comfort}}  \\ \hline
{\texttt{beautiful}, \texttt{lovely}, \texttt{gorgeous}, \texttt{handsome}, \texttt{beautifully}}  \\ \hline
{\texttt{guy}, \texttt{guys}, \texttt{Guy}, \texttt{dude}}  \\ \hline
{\texttt{fit}, \texttt{FIT}, \texttt{fits}, \texttt{fitting}, \texttt{fitted}}  \\ \hline
{\texttt{recomm}, \texttt{recommend}, \texttt{recommended}, \texttt{recommendation}}  \\ \hline
{\texttt{star}, \texttt{stars}, \texttt{Stars}}  \\ \hline
{\texttt{reach}, \texttt{reached}, \texttt{reaching}, \texttt{reaches}, \texttt{\#reach}}  \\ 
\bottomrule
\end{tabular}}
\caption{Cluster examples of \ours-8000. Each row presents tokens belonging to a single cluster.}\label{tab:cluster_examples_2}
\end{table}

\subsection{Training data details}\label{data_details}
We leverage the public training data provided by BGE-en-ICL~\citep{bge_en_icl}. Specifically, the training data is a mixture of retrieval, reranking, classification, clustering, and STS data. 
\begin{itemize}
    \item \textbf{Retrieval}:  ELI5~\citep{fan2019eli5}, HotpotQA~\citep{yang2018hotpotqa}, FEVER~\citep{thorne2018fever}, MSMARCO passage and document ranking~\citep{msmarco}, NQ, NLI, SQuAD, TriviaQA, Quora Duplicate Questions~\citep{quora-question-pairs}, Arguana \citep{wachsmuth2018retrieval}, FiQA \citep{maia201818}.

    \item \textbf{Reranking}: SciDocsRR \citep{cohan2020specter}, StackOverFlowDupQuestions \citep{liu2018linkso}.

    \item \textbf{Classification}: AmazonReviews-Classification \citep{mcauley2013hidden}, AmazonCounterfactual-Classification \citep{o2021wish}, Banking77-Classification \citep{casanueva2020efficient}, Emotion-Classification \citep{saravia2018carer}, TweetSentimentExtraction-Classification \citep{tweet-sentiment-extraction}, MTOPIntent-Classification \citep{li2020mtop}, IMDB-Classification \citep{maas2011learning}, ToxicConversations-Classification \citep{kaggle2019jigsaw}.

    \item \textbf{Clustering}: TwentyNewsgroups-Clustering \citep{lang1995newsweeder}, \{Arxiv/Biorxiv/Medrxiv/Reddit/StackExchange\}-Clustering-\{S2S/P2P\}

    \item \textbf{STS}: STS12 \citep{agirre2012semeval}, STS22 \citep{chen2022semeval}, STS-Benchmark \citep{cer2017semeval}.
\end{itemize}

\subsection{Task Instructions}\label{sec:instructions}
We present the task instructions we used in Table~\ref{table:evalinstructions}.
\begin{table*}[h]
\begin{center}
\begin{small}
\setlength{\tabcolsep}{4.5pt}
\setlength{\extrarowheight}{2pt}
\resizebox{\textwidth}{!}{
\begin{tabular}{ll}
\toprule
Task Name & Instruction \\ \midrule
ArguAna    &     Given a claim, find documents that refute the claim. \\
ClimateFEVER    &     \begin{tabular}[c]{@{}l@{}}Given a claim about climate change, retrieve documents that support or refute \\ the claim.\end{tabular} \\
CQADupStack    &   \begin{tabular}[c]{@{}l@{}}Given a question, retrieve detailed question descriptions from Stackexchange that are \\ duplicates to the given question.\end{tabular} \\
DBPedia    &     Given a query, retrieve relevant entity descriptions from DBPedia. \\
FEVER    &     Given a claim, retrieve documents that support or refute the claim. \\
FiQA2018    &     Given a financial question, retrieve user replies that best answer the question. \\
HotpotQA    &     Given a multi-hop question, retrieve documents that can help answer the question. \\
MSMARCO    &     Given a web search query, retrieve relevant passages that answer the query. \\
NFCorpus    &     Given a question, retrieve relevant documents that best answer the question. \\
Natural Question    &     Given a question, retrieve Wikipedia passages that answer the question. \\
QuoraRetrieval    &     \begin{tabular}[c]{@{}l@{}}Given a question, retrieve questions that are semantically equivalent to the given \\ question.\end{tabular} \\
SCIDOCS    &     Given a scientific paper title, retrieve paper abstracts that are cited by the given paper. \\
SciFact    &     Given a scientific claim, retrieve documents that support or refute the claim. \\
Touche2020    &     Given a question, retrieve detailed and persuasive arguments that answer the question. \\
TREC-COVID    &     Given a query, retrieve documents that answer the query. \\ 
STS*    &     Retrieve semantically similar text. \\
SummEval    &     Given a news summary, retrieve other semantically similar summaries. \\
AmazonCounterfactualClassification    &     \begin{tabular}[c]{@{}l@{}}Classify a given Amazon customer review text as either counterfactual \\ or not-counterfactual.\end{tabular} \\
AmazonPolarityClassification    &     Classify Amazon reviews into positive or negative sentiment. \\
AmazonReviewsClassification    &     Classify the given Amazon review into its appropriate rating category. \\
Banking77Classification    &     Given a online banking query, find the corresponding intents. \\
EmotionClassification    &   \begin{tabular}[c]{@{}l@{}}Classify the emotion expressed in the given Twitter message into one of the six \\ emotions: anger, fear, joy, love, sadness, and surprise.\end{tabular}   \\ 
ImdbClassification    &     \begin{tabular}[c]{@{}l@{}}Classify the sentiment expressed in the given movie review text from \\ the IMDB dataset.\end{tabular} \\
MassiveIntentClassification    &     Given a user utterance as query, find the user intents. \\
MassiveScenarioClassification    &     Given a user utterance as query, find the user scenarios. \\
MTOPDomainClassification    &     Classify the intent domain of the given utterance in task-oriented conversation. \\
MTOPIntentClassification    &     Classify the intent of the given utterance in task-oriented conversation. \\
ToxicConversationsClassification    &     Classify the given comments as either toxic or not toxic. \\
TweetSentimentExtractionClassification    &     Classify the sentiment of a given tweet as either positive, negative, or neutral. \\
ArxivClusteringP2P    &     \begin{tabular}[c]{@{}l@{}}Identify the main and secondary category of Arxiv papers based on the titles \\ and abstracts.\end{tabular} \\
ArxivClusteringS2S    &     Identify the main and secondary category of Arxiv papers based on the titles. \\
BiorxivClusteringP2P    &     Identify the main category of Biorxiv papers based on the titles and abstracts. \\
BiorxivClusteringS2S    &     Identify the main category of Biorxiv papers based on the titles. \\
MedrxivClusteringP2P    &     Identify the main category of Medrxiv papers based on the titles and abstracts. \\
MedrxivClusteringS2S    &     Identify the main category of Medrxiv papers based on the titles. \\
RedditClustering    &     Identify the topic or theme of Reddit posts based on the titles. \\
RedditClusteringP2P    &     Identify the topic or theme of Reddit posts based on the titles and posts. \\
StackExchangeClustering    &     Identify the topic or theme of StackExchange posts based on the titles. \\
StackExchangeClusteringP2P    &     Identify the topic or theme of StackExchange posts based on the given paragraphs. \\
TwentyNewsgroupsClustering    &     Identify the topic or theme of the given news articles. \\
AskUbuntuDupQuestions    &     Retrieve duplicate questions from AskUbuntu forum. \\
MindSmallReranking    &     Retrieve relevant news articles based on user browsing history. \\
SciDocsRR    &     Given a title of a scientific paper, retrieve the titles of other relevant papers. \\
StackOverflowDupQuestions    &     Retrieve duplicate questions from StackOverflow forum. \\
SprintDuplicateQuestions    &     Retrieve duplicate questions from Sprint forum. \\
TwitterSemEval2015    &     Retrieve tweets that are semantically similar to the given tweet. \\
TwitterURLCorpus    &     Retrieve tweets that are semantically similar to the given tweet. \\ \midrule
AIR-Bench    &     Given a question, retrieve passages that answer the question. \\ \bottomrule
\end{tabular}
}
\end{small}
\end{center}
\caption{Task instructions for MTEB and AIR-Bench benchmarks.}
\label{table:evalinstructions}
\end{table*}

\subsection{MTEB Subset Details}\label{sec:subset_details}
Following~\citet{repurposing}, for each task category, we select one dataset for evaluation. The chosen dataset is determined based on the model results presented in the original MTEB paper, focusing on the dataset with the highest correlation to the category's average performance.
\begin{itemize}
    \item \textbf{Classification}: EmotionClassification
    \item \textbf{Clustering}: TwentyNewsgroupsClustering
    \item \textbf{Pair classification}: SprintDuplicateQuestions
    \item \textbf{Reranking}: AskUbuntuDupQuestions
    \item \textbf{Retrieval}: SciFact
    \item \textbf{Semantic text similarity}: STS15
    \item \textbf{Summarization}: SummEval
\end{itemize} 

\subsection{Detailed MTEB Results}
We present the detailed MTEB results in Table~\ref{table:fullmteb}.

\begin{table*}[h]
\begin{center}
\hspace*{-1.6cm}
\begin{small}\begin{tabular}{l|ccc|ccccc}
\toprule
Dataset & \begin{tabular}[c]{@{}l@{}}gte-Qwen2-\\ 7B-instruct\end{tabular} & \begin{tabular}[c]{@{}l@{}}SFR-Embe\\ dding-2\_R\end{tabular} & \begin{tabular}[c]{@{}l@{}}stella\_en\_\\ 1.5B\_v5\end{tabular} & \begin{tabular}[c]{@{}l@{}}BGE-en-ICL\\ (zero-shot)\end{tabular} & \begin{tabular}[c]{@{}l@{}}NV-Em\\ bed-v2\end{tabular} & \begin{tabular}[c]{@{}l@{}}\ours\\-4000\end{tabular} & \begin{tabular}[c]{@{}l@{}}\ours\\-8000\end{tabular} \\ \hline
ArguAna & 64.27 & 62.34 & 65.27 & 82.76 & 70.07 & 77.32 & 76.02  \\
ClimateFEVER & 45.88 & 34.43 & 46.11 & 45.35 & 45.39 & 44.62 & 45.77   \\
CQADupStack & 46.43 & 46.11 & 47.75 & 47.23 & 50.24 & 47.39 & 48.67  \\
DBPEDIA & 52.42 & 51.21 & 52.28 & 50.42 & 53.50 & 50.10 & 49.75  \\
FEVER & 95.11 & 92.16 & 94.83 & 91.96 & 93.75 & 92.37 & 92.32  \\
FiQA2018 & 62.03 & 61.77 & 60.48 & 58.77 & 65.73 & 60.43 & 61.57  \\
HotpotQA & 73.08 & 81.36 & 76.67 & 84.98 & 85.48 & 85.07 & 85.71  \\
MSMARCO & 45.98 & 42.18 & 45.22 & 46.72 & 45.63 & 46.95 & 47.24  \\
NFCorpus & 40.60 & 41.34 & 42.00 & 40.69 & 45.17 & 41.64 & 40.61  \\
Natural Question & 67.00 & 73.96 & 71.80 & 73.85 & 73.57 & 73.13 & 74.64   \\
QuoraRetrieval & 90.09 & 89.58 & 90.03 & 91.02 & 89.04 & 90.84 & 90.79  \\
SCIDOCS & 28.91 & 24.87 & 26.64 & 25.25 & 21.90 & 27.51 & 28.54  \\
SciFact & 79.06 & 85.91 & 80.09 & 78.33 & 80.13 & 78.39 & 79.75 \\
Touche2020 & 30.57 & 28.18 & 29.94 & 29.67 & 31.78 & 25.86 & 29.34  \\
TREC-COVID & 82.26 & 87.28 & 85.98 & 78.11 & 88.44 & 69.73 & 77.18  \\
BIOSSES & 81.37 & 87.60 & 83.11 & 86.35 & 87.42 & 84.47 & 85.83  \\
SICK-R & 79.28 & 77.01 & 82.89 & 83.87 & 82.15 & 83.81 & 83.30  \\
STS12 & 79.55 & 75.67 & 80.09 & 77.73 & 77.89 & 79.07 & 80.99  \\
STS13 & 88.83 & 82.40 & 89.68 & 85.98 & 88.30 & 86.54 & 87.34  \\
STS14 & 83.87 & 79.93 & 85.07 & 82.34 & 84.30 & 84.32 & 84.39  \\
STS15 & 88.54 & 85.82 & 89.39 & 87.35 & 89.04 & 89.69 & 89.75 \\
STS16 & 86.49 & 84.50 & 87.15 & 86.54 & 86.77 & 87.23 & 87.63  \\
STS17 & 88.73 & 88.93 & 91.35 & 91.25 & 90.67 & 91.55 & 90.87  \\
STS22 & 66.88 & 67.10 & 68.10 & 68.08 & 68.12 & 68.69 & 68.09  \\
STSBenchmark & 86.85 & 83.60 & 88.23 & 87.92 & 88.41 & 88.22 & 88.47  \\
SummEval & 31.35 & 30.71 & 31.49 & 30.75 & 30.70 & 31.55 & 29.54  \\
SprintDuplicateQuestions & 92.82 & 97.62 & 96.04 & 95.06 & 97.02 & 96.98 & 97.00   \\
TwitterSemEval2015 & 77.96 & 78.57 & 80.58 & 78.54 & 81.11 & 79.31 &  79.56 \\
TwitterURLCorpus & 86.59 & 88.03 & 87.58 & 87.19 & 87.87 & 87.50 &  87.37 \\
AmazonCounterfactual & 91.31 & 92.72 & 92.87 & 92.88 & 94.28 & 93.61 & 93.69  \\
AmazonPolarity & 97.50 & 97.31 & 97.16 & 96.86 & 97.74 & 97.05 & 97.07  \\
AmazonReviews & 62.56 & 61.04 & 59.36 & 61.28 & 63.96 & 62.83 & 63.61  \\
Banking77 & 87.57 & 90.02 & 89.79 & 91.42 & 92.42 & 90.43 &  90.19 \\
Emotion & 79.45 & 93.37 & 84.29 & 93.31 & 93.38 & 92.33 & 91.87  \\
Imdb & 96.75 & 96.80 & 96.66 & 96.91 & 97.14 & 97.12 & 97.00  \\
MassiveIntent & 85.41 & 85.97 & 85.83 & 82.26 & 86.10 & 79.65 & 81.14  \\
MassiveScenario & 89.77 & 90.61 & 90.20 & 83.92 & 92.17 & 81.97 & 83.53  \\
MTOPDomain & 99.04 & 98.58 & 99.01 & 97.99 & 99.25 & 97.49 & 97.44 \\
MTOPIntent & 91.88 & 91.30 & 92.78 & 93.56 & 94.37 & 92.59 & 92.81\\
ToxicConversations & 85.12 & 91.14 & 88.76 & 93.16 & 92.74 & 92.29 & 92.37 \\
TweetSentimentExtraction & 72.58 & 79.70 & 74.84 & 79.90 & 80.87 & 80.17 & 80.42 \\
Arxiv-P2P & 54.46 & 54.02 & 55.44 & 54.42 & 55.80 & 54.87 & 54.81 \\
Arxiv-S2S & 51.74 & 48.82 & 50.66 & 49.17 & 51.26 & 50.25 & 50.14 \\
Biorxiv-P2P & 50.09 & 50.76 & 50.68 & 52.32 & 54.09 & 52.39 & 52.48 \\
Biorxiv-S2S & 46.65 & 46.57 & 46.87 & 48.38 & 49.60 & 48.35 & 48.52 \\
Medrxiv-P2P & 46.23 & 46.66 & 46.87 & 46.13 & 46.09 & 46.35 & 46.38 \\
Medrxiv-S2S & 44.13 & 44.18 & 44.65 & 44.20 & 44.86 & 44.54 & 44.89 \\
Reddit & 73.55 & 62.92 & 72.86 & 71.20 & 71.10 & 72.32 & 72.37 \\
Reddit-P2P & 74.13 & 72.74 & 75.27 & 72.17 & 74.94 & 73.20 & 73.89 \\
StackExchange & 79.86 & 76.48 & 80.29 & 81.29 & 82.10 & 81.70 & 81.60 \\
StackExchange-P2P & 49.41 & 48.29 & 49.57 & 45.53 & 48.36 & 43.73 & 44.41 \\
TwentyNewsgroups & 53.91 & 66.42 & 61.43 & 68.51 & 64.82 & 69.44 & 68.78 \\
AskUbuntuDupQuestions & 67.58 & 66.71 & 67.33 & 64.80 & 67.46 & 65.45 & 65.74 \\
MindSmallRerank & 33.36 & 31.26 & 33.05 & 30.60 & 31.76 & 31.92 & 31.46 \\
SciDocsRR & 89.09 & 87.29 & 89.20 & 86.90 & 87.59 & 87.92 & 87.63 \\
StackOverflowDupQuestions & 55.66 & 55.32 & 55.25 & 56.32 & 55.79 & 58.15 & 58.79 \\ \hline
\bf{MTEB Average (56)} & 70.24 & 70.31 & 71.19 & 71.24 & 72.31 & 71.22 & 71.63 \\ 
\bottomrule
\end{tabular}
\end{small}
\end{center}
\vspace{-0.4cm}
\caption{Detailed MTEB results.}
\label{table:fullmteb}
\end{table*}

\end{document}